\documentclass{article} 
\usepackage{iclr2024_conference,times}

\usepackage{amsmath,amsfonts,bm}

\def\eqref#1{equation~\ref{#1}}

\def\1{\bm{1}}

\DeclareMathAlphabet{\mathsfit}{\encodingdefault}{\sfdefault}{m}{sl}
\SetMathAlphabet{\mathsfit}{bold}{\encodingdefault}{\sfdefault}{bx}{n}

\usepackage{hyperref}
\usepackage{url}
\usepackage{graphicx}
\usepackage{xspace,mfirstuc,tabulary}
\usepackage{times}
\usepackage{latexsym}
\usepackage{booktabs}
\usepackage{multirow}
\usepackage{comment}
\usepackage{dirtytalk}
\usepackage{color, colortbl}
\usepackage{adjustbox}
\usepackage{todonotes}
\usepackage{lingmacros}
\usepackage{lipsum} 
\usepackage{subcaption}
\usepackage{cleveref}

\newcommand*{\yoruba}{Yor\`ub\'a\xspace}
\newcommand*{\menyo}{MENYO-20k\xspace}

\definecolor{GrayColor}{gray}{0.9}

\title{YAD: Leveraging T5 for improved automatic diacritization of \yoruba text}

\author{Akindele Michael Olawole\textsuperscript{1}\thanks{Equal contribution.}, Jesujoba O. Alabi\textsuperscript{2*}, Aderonke Busayo Sakpere\textsuperscript{1}, David I. Adelani\textsuperscript{3} \\
\textsuperscript{1} Department of Computer Science, University of Ibadan, Nigeria  \\
\textsuperscript{2} Spoken Language Systems (LSV), Saarland University,
Saarland Informatics Campus, Germany\\
\textsuperscript{3} University College London \\
\texttt{jalabi@lsv.uni-saarland.de}, \texttt{d.adelani@ucl.ac.uk} \\
}

\iclrfinalcopy 
\begin{document}

\maketitle

\begin{abstract}
In this work, we present Yorùbá automatic diacritization (YAD) benchmark dataset for evaluating \yoruba diacritization systems. In addition, we pre-train text-to-text transformer, T5 model for \yoruba and showed that this model outperform several multilingually trained T5 models. Lastly, we showed that more data and larger models are better at diacritization for \yoruba
\end{abstract}

\paragraph{Introduction}
\yoruba, a language spoken predominantly in West Africa, is renowned for its tonal nature which is characterized by a heavy use of diacritics to signify tone variations. In \yoruba and many other languages, diacritics play a crucial role in disambiguating word meanings and in word pronunciation, making accurate diacritization essential for effective communication and language processing tasks~\citep{skiredj2024arabic}. However, manual diacritization is time-consuming and requires specialized linguistic expertise, motivating the development of automatic diacritization systems. In recent years, significant progress has been made in natural language processing (NLP) techniques, leading to the exploration of various approaches to automate the diacritization process for languages using diacritics~\cite[\textit{inter alia}]{naplava-etal-2018-diacritics, mubarak-etal-2019-highly, naplava-etal-2021-diacritics, stankevicius-etal-2022-correcting} including \yoruba~\citep{orife2018attentive, orife2020improving}. Despite these efforts, there is the absence of a standard benchmark dataset for evaluating and comparing developed \yoruba diacritization systems. Furthermore, the emphasis has predominantly been on model development rather than assessing the usability of these systems, particularly in the context of lightweight models amidst the prevalence of large neural networks in contemporary NLP research. In this study, we address these gaps by introducing YAD (\yoruba Automatic Diacritization Dataset), which we curated by leveraging the \menyo~\citep{adelani-etal-2021-effect} dataset. Moreover, we focus on developing lightweight diacritizers for Yoruba, leveraging T5~\citep{2020t5} a text-to-text transformer architecture, to improve usability and efficiency in the era of large neural networks. For reproducibility, we release our code and data on GitHub.\footnote{\url{https://github.com/ajesujoba/YAD}}

\paragraph{\yoruba language} \yoruba orthography is based on the Latin alphabet. Its alphabets include 25 Latin letters excluding characters c, q, v, x and z, and its has five additional letter ({\d e}, gb, {\d s} , {\d o}) representing certain sounds. Diacritics including accents and undots are used in \yoruba to capture tone and vowel height. The tones include low, middle and high which are denoted by the grave (e.g. ``\`{a}''), optional macron (e.g. ``\={a}'') and acute (e.g. ``\'{a}'') accents respectively. 

\paragraph{\yoruba automatic diacritization dataset (YAD)} 
In this work, we present YAD, a dataset created for evaluating diacritization systems for \yoruba. YAD is constructed based on the \menyo dataset, which is a benchmark for English-\yoruba translation, containing 20,000 parallel sentences. We divided the \yoruba side of \menyo's test data into two halves, allocating one half as development set (Dev.) and the other half as test set (Test). We ensured that both halves of the data had equal representation across all domains present in \menyo. Given these splits, we applied four heuristics on the data, considering the sequence-to-sequence nature of the diacritization task. This task requires a source side and a target side, where the source side can consist of \yoruba text without diacritics or partially diacritized texts, and the target side is fully diacritized \yoruba texts. To create and prepare the source side of the data, the following heuristics were applied with specific proportions on the original, well-diacritized texts: (i) removal of all diacritics for $60\%$ of the data, (ii) removal of only tone marks for $20\%$ of the data, (iii) a combination of (ii) and the removal of any random word for $10\%$, and (iv) a combination of (ii) and swapping any two distinct words (applicable only for sentences with more than two words). Where (iii) and (iv) can be seen as infilling and grammatical error correction tasks respectively. Example~\ref{ex:example1}, shows an example from the development set where (iv) was applied where sib{\d e}sib{\d e}, and {\d S}ugb{\d o}n were swapped on the source side.

\enumsentence{\label{ex:example1}\small\textbf{source}: sib{\d e}sib{\d e}, {\d S}ugb{\d o}n Mama o gbagb{\d o}. \\
\textbf{reference}: {\d S}{\` u}gb{\d {\' u}}n s{\' i}b{\d {\` e}}s{\' i}b{\d {\` e}}, M{\` a}m{\' a} {\` o} gb{\` a}gb{\d {\' o}}}

\paragraph{Diacritization model} We pre-train T5 models from scratch using \yoruba WURA corpus~\citep{oladipo-etal-2023-better} and JW300~\citep{agic-vulic-2019-jw300} with different sizes: tiny (number of layers encoder/decoder (L=4), heads (H=4)) , mini (L=4, H=4), small (L=8, H=6), base (L=12, H=12) corresponding to 14M, 18M, 60M and 280M parameters. All models make use of a vocabulary size of 32K except the tiny model. We called our model o\textbf{yo}-T5.~\footnote{Oyo \yoruba (YO) dialect of T5. Oyo dialect is the standardized dialect for writing \yoruba}. All models were pre-trained using HuggingFace Flax code and TPU v3-8 for 100-200K steps, which took around one day each. 

After pre-training, we trained our baseline diacritizer models by fine-tuning existing multilingually trained language models, MT5~\citep{xue-etal-2021-mt5}, AfriMT5~\citep{adelani-etal-2022-thousand}, AfriTeVa-V2~\citep{oladipo-etal-2023-better}, and UMT5~\citep{chung2023unimax}. We use MT5-base,  Afri-MT5-base, the base and large variants of AfriTeVa-V2, and UMT5. we fine-tune these models using HuggingFace Transformers~\citep{wolf-etal-2020-transformers}. These eventual diacritization models are evaluated using the test and development SacreBLEU~\citep{post-2018-call}\footnote{``intl'' tokenizer, all data comes untokenized.} and ChrF~\citep{popovic-2015-chrf}.

\paragraph{Training and evaluation data} For our experiments, we trained diacritization models using data from three sources, which are, the \yoruba Bible, combination of training and development split of \menyo (now referred to as YAD training split from here onwards), and JW300~\citep{agic-vulic-2019-jw300}. \Cref{tab:data_count} shows the number of sentences in each split. Different combinations of these datasets were used for training our models, and unless specified otherwise we implemented the four previously described heuristics on the training data also. We evaluated our models on the YAD development and test splits, as well as on an existing Global Voices (GV) test set from~\cite{orife2020improving}. Additionally, we used a sampled $10\%$ of the Bible as a test set. For reproducibility, we release our code, data and models on GitHub\footnote{\url{https://github.com/ajesujoba/YAD}}.

\paragraph{Baseline  models and result} We trained diacritization models by fine-tuning MT5, Afri-MT5, AfriTeVa-V2, and UMT5, on the YAD training split. The experimental results as presented in \Cref{tab:main_result} shows that AfriTeVa-V2-large outperform other massively multilingual language models. Interestingly, Oyo-T5-base demonstrates competitiveness on both BLEU and CHRF similar to AfriTeVa-V2-large, despite having fewer parameters compared to AfriTeVa-V2-large. 

\begin{table}[htbp]
\scalebox{0.8}{%
\begin{minipage}[b]{0.4\linewidth}
    \centering
    \begin{tabular}{cccc}
        \toprule
        \textbf{Corpus} & \textbf{Train} & \textbf{Dev} & \textbf{Test} \\
        \midrule 
        Bible & 27739 & - & 3083 \\
        JW300 & 459,871 & - & - \\ 
        \rowcolor{GrayColor}
        YAD & 13,467 & 3,326 & 3,330 \\
        \midrule
        \multicolumn{3}{l}{Previous test set: \cite{orife2020improving}} \\
        \rowcolor{GrayColor}
        GV & - & - & 619 \\
        \bottomrule
    \end{tabular}
    \caption{Data split of different corpus.}
    \label{tab:data_count}
\end{minipage}}%
\scalebox{0.7}{%
\begin{minipage}[b]{1.0\linewidth}
    \centering
    \begin{tabular}{l|c|rrrr}
        \toprule
         &  & \multicolumn{2}{c}{\textbf{Dev. eval.}} & \multicolumn{2}{c}{\textbf{Test eval.}} \\
        \textbf{Model} & \textbf{Size} & \textbf{BLEU} & \textbf{CHRF} & \textbf{BLEU} & \textbf{CHRF} \\
        \midrule
        UMT5-base & 580M & 38.1 & 62.5 & 39.3 & 64.0 \\
        MT5-base & 580M & 48.1 & 69.4 & 49.5 & 71.2 \\
        AfriMT5-base & 580M & 53.3 & 72.8 & 54.0 & 74.2 \\
        AfriTeVa-base & 313M & 56.9 & 72.9 & 57.2 & 74.0 \\
        AfriTeVa-large & 871M & 67.2 & 80.8 & 66.8 & 81.3 \\
        Oyo-T5-base & 280M & \underline{\textbf{71.9}} & \underline{\textbf{82.2}} & \underline{\textbf{70.2}} & \underline{\textbf{82.3}} \\

        \bottomrule
    \end{tabular}
    \caption{Diacritization model evaluation on YAD Dev. and Test sets.}
    \label{tab:main_result}
\end{minipage}}
\end{table}

\paragraph{Effect of model scaling} Given that Oyo-T5-base is a competitive model despite its size, we chose to train light-weight diacritizers by finetuning the smaller variants of Oyo-T5 model on the YAD training data to check the effect of model size on this task. We evaluated the models on YAD dev. and test sets. As shown in \Cref{tab:model_scale}, bigger models are better for this task. However, Oyo-T5-small with 60M parameters even outperformed AfriTeVa-base with 313M parameters on the same task.

\paragraph{Effect of training data scaling} Going further, we choose to evaluate the effect of training data size on the diacritization models. Hence trained new diacritization models on Bible, JW300, and a combination of them and YAD training data using Oyo-T5-base as the backbone model and we evaluate them on the Bible, GV, and YAD test sets. For these test sets, we use the version where all diacritics were removed from the source side without applying the earlier described heuristics \footnote{We also analyze this effect on text with only underdot on the source side and provide the result in \Cref{sec:inputside}}.

\Cref{tab:data_scale} shows that training on more data is better. Also training on just JW300 outperformed the best result achieved by \citet{orife2020improving} on GV by $+11.6$ BLEU point. Furthermore, we observed that models trained on a specific domain tend to perform better on that domain. For instance, models trained on the Bible achieved a $+17.6$ BLEU point improvement compared to the next best performing model, which was trained on JW300. It is important to note that YAD's training data, obtained from the combination of \menyo training and development splits, included Global Voices news. Therefore, we suspect that its high BLEU scores on the GV tests set could be attributed to potential data leakage. Lastly, our findings suggest that more data generally leads to better performance, as evidenced by the results from the model trained on the combined dataset.

\begin{table}[htbp]
\begin{minipage}[t]{0.4\linewidth}
\centering

\scalebox{0.8}{%
\begin{tabular}{l|ccc}
\toprule
\textbf{Model/Data} & \textbf{Bible} & \textbf{GV} & \textbf{YAD} \\
\midrule
\multicolumn{4}{l}{Data: JW300+Bible+others} \\
\cite{orife2020improving}  & - & 59.8 & - \\
\midrule
Bible & 88.1 & 55.0 & 57.2 \\
JW300 & 71.7 & 71.4  & 66.9   \\
YAD & 69.0 & 82.1 & 72.0  \\
JW300+Bible+YAD & 90.8 & 91.0 & \textbf{78.7} \\ 
\bottomrule
\end{tabular}
}
\caption{Data scale effect on Bible, GV \& YAD.}
\label{tab:data_scale}
\end{minipage}%
\begin{minipage}[t]{0.45\linewidth}
\centering
\scalebox{0.89}{%
\begin{tabular}{l|c|cc|cc}
\toprule
 & & \multicolumn{2}{c|}{\textbf{Dev. eval.}} & \multicolumn{2}{c}{\textbf{Test eval.}} \\
\midrule
\textbf{Model} & \textbf{Size} & \textbf{BLEU} & \textbf{CHRF} & \textbf{BLEU} & \textbf{CHRF} \\
\midrule
Oyo-T5-tiny & 14M & 42.1 & 62.1 & 36.3 & 60.4 \\
Oyo-T5-mini & 18M & 49.0 & 64.6 & 47.5 & 65.6 \\
Oyo-T5-small & 60M & 63.5 & 75.8 & 62.9 & 76.7 \\
Oyo-T5-base & 280M & \underline{\textbf{71.9}} & \underline{\textbf{82.2}} & \underline{\textbf{70.2}} & \underline{\textbf{82.3}} \\ 
\bottomrule
\end{tabular}
}
\caption{Effect of the model scale on YAD.\looseness-1}
\label{tab:model_scale}
\end{minipage}
\end{table}

\paragraph{Conclusion} In this work we present \yoruba automatic diacritization (YAD) benchmark dataset for evaluating \yoruba diacritization systems. In addition, we pre-train text-to-text transformer, T5 model for \yoruba and showed that this model outperform several multilingually trained T5 models. Lastly, we showed that more data and bigger models are better at diacritization for \yoruba.

\subsubsection*{Acknowledgments}
Jesujoba Alabi was partially supported by the BMBF’s (German Federal Ministry of Education and Research) SLIK project under the grant 01IS22015C. David Adelani acknowledges the support of DeepMind Academic Fellowship programme. This work was supported in part by Oracle Cloud credits and related resources provided by Oracle.

\bibliography{iclr2024_conference}
\bibliographystyle{iclr2024_conference}

\appendix
\section{Effect of applying diacritics to undiacritized texts versus text with only underdot}
\label{sec:inputside}
\begin{table}[h]
\centering
\scalebox{0.9}{%
\begin{tabular}{l|cc|cc|cc}
\toprule
 & \multicolumn{2}{c|}{\textbf{Bible}} & \multicolumn{2}{c|}{\textbf{GV.}} & \multicolumn{2}{c}{\textbf{YAD.}}\\
\midrule
\textbf{Model} & \textbf{BLEU} & \textbf{CHRF} & \textbf{BLEU} & \textbf{CHRF}& \textbf{BLEU} & \textbf{CHRF} \\
\midrule
\multicolumn{7}{c}{\textbf{no accent on source side}} \\
Bible &  88.2 & 94.3 & 55.2 & 68.4 & 57.8 & 72.9 \\
JW300 &  64.0 & 80.9 & 68.8 & 81.6 & 65.0 & 80.9 \\
Menyo &  71.8 & 83.9 & 79.3 & 86.7 & 72.0 & 83.7 \\
JW300+Bible+Menyo & 89.6 & 95.8 & 90.0 & 95.1 & 78.4 & 89.0 \\ 
\midrule
\multicolumn{7}{c}{\textbf{no diacritics on source side}} \\
Bible &  88.1 & 94.2 & 55.0 & 67.4 & 57.2 & 71.6 \\
JW300 &  71.7 & 82.6 & 71.4 & 83.7 & 66.9 & 81.9 \\
Menyo &  69.0 & 82.4 & 82.1 & 87.3 & 72.0 & 83.3 \\
JW300+Bible+Menyo & 90.8 & 96.0 & 91.0 & 95.1 & 78.7 & 88.8 \\ 
\bottomrule
\end{tabular}
}
\caption{Effect of applying diacritics to undiacritized texts vs text with only underdots}
\label{tab:model_scale2}
\end{table}

\end{document}